\documentclass[times,twocolumn,final]{elsarticle}

\usepackage{framed,multirow}

\usepackage{amssymb}
\usepackage{latexsym}
\usepackage{lineno,hyperref}
\modulolinenumbers[5]
\usepackage{framed,multirow}
\usepackage{float}
\usepackage{booktabs}
\usepackage{placeins}
\usepackage[ruled,vlined]{algorithm2e}
\usepackage{setspace}
\usepackage{blindtext}
\usepackage{multicol}
\usepackage{comment}

\usepackage{multirow}

\usepackage{url}
\usepackage{xcolor}
\usepackage{float}
\usepackage{blindtext}
\usepackage{caption}

\usepackage{hyperref}

\usepackage{soul}
\usepackage{todonotes}

\definecolor{newcolor}{rgb}{.8,.349,.1}

\journal{Medical Image Analysis}

\begin{document}



\begin{frontmatter}

\title{Texture Characterization of Histopathologic Images Using Ecological Diversity Measures and Discrete Wavelet Transform}
\tnotetext[tnote1]{This is an example for title footnote coding.}

\author[1]{Steve Tsham Mpinda {Ataky}}
\cortext[cor1]{Corresponding author: 
  Tel.: +0-000-000-0000;  
  fax: +0-000-000-0000;}
\author[1]{Alessandro  {Lameiras Koerich}\corref{cor1}}
\fntext[fn1]{This is author footnote for second author.}

\address[1]{École de Technologie Supérieure, Université du Québec,  1100 Notre-Dame St W, Montreal, Quebec H3C 1K3, Canada}


\begin{abstract}
Breast cancer is a health problem that affects mainly the female population. An early detection increases the chances of effective treatment, improving the prognosis of the disease. In this regard, computational tools have been proposed to assist the specialist in interpreting the breast digital image exam, providing features for detecting and diagnosing tumors and cancerous cells. Nonetheless, detecting tumors with a high sensitivity rate and reducing the false positives rate is still challenging. Texture descriptors have been quite popular in medical image analysis, particularly in histopathologic images (HI), due to the variability of both the texture found in such images and the tissue appearance due to irregularity in the staining process. Such variability may exist depending on differences in staining protocol such as fixation, inconsistency in the staining condition, and reagents, either between laboratories or in the same laboratory. Textural feature extraction for quantifying HI information in a discriminant way is challenging given the distribution of intrinsic properties of such images forms a non-deterministic complex system. This paper proposes a method for characterizing texture across HIs with a considerable success rate. By employing ecological diversity measures and discrete wavelet transform, it is possible to quantify the intrinsic properties of such images with promising accuracy on two HI datasets compared with state-of-the-art methods.
\end{abstract}


\end{frontmatter}



\section{Introduction}
\label{sec:intro}
Cancer is a disease caused by an uncontrolled division of cells that can happen anywhere in the body, susceptible to invading neighboring tissues and organs ~\citep{ferreira:04}. Cell proliferation does not necessarily imply the presence of malignancy but can simply respond to the body's specific needs. More than a hundred known types of tumors affect people, but not all tumors are cancerous ~\citep{nci:16}. Benign tumors grow in an organized, generally slow, expansive manner and have apparent limits. Although they do not invade neighboring tissues, they can compress adjacent organs and tissues and are hardly life-threatening. In contrast, malignant tumors manifest a greater degree of autonomy. They are capable of invading neighboring tissues and causing metastases, which may be resistant to treatment and cause the patient's death ~\citep{instituto:11}. 

Breast cancer is the most frequent cancer among women globally, accounting for nearly a quarter of new cases each year ~\citep{inca:18}, and is the most prevalent type of cancer in nearly 140 countries~\citep{Torre2017}. In addition to resulting in death, related treatment costs are high. It impacts both the public and the private healthcare system, which, in turn, penalizes both the government and the population. 

Digital mammography analysis is among the most widely employed and effective methods for the early detection of breast cancer. Mammography is a radiological test that produces images in gray-scale, which are analyzed by radiologists, who need to give more attention and spend more time reliably detecting the cancer information from the images \citep{GIGER:2000}. Likewise, imaging examinations like computerized tomography or ultrasound can diagnose whether there are masses growing in breast tissue, despite the verification of either type of tumor can only be accomplished employing biopsy. Biopsies, in turn, last to provide a result because of the acquisition procedure, which may imply open surgical biopsy or fine-needle aspiration, tissue processing (preparation of slides with a staining process) as well as pathologist visual examination. Innately, pathologist examination is a time-consuming and highly specialized task prone to intra and inter-observer disagreement~\citep{BELLOCQ2011S92}. 

Histopathology is the study of how a specific disease affects a set of cells (tissue). Usually, a biopsy study is done using a microscope and dyes. It can also be done during surgery or at an autopsy (death investigation). Histopathologic images (HIs) are a type of medical imaging obtained via microscopy of tissues from biopsies, which allow specialists to observe tissue characteristics on a cell basis. This process consists of tissue processing by chemical fixation or frozen section slides (creating the slide with the staining process). Next, the slides undergo staining process with one or several pigments to envision the tissue by means of a microscope, aiming to reveal cellular components; counterstains are used to provide contrast, and finally, the pathologist analysis. The stain by and large used in histopathology is a combination of hematoxylin and eosin (H\&E). The former is employed to stain nuclei (blue), while the latter stains cytoplasm and the extracellular connective tissue matrix (pink). However, the staining process may give rise to a variance in the analysis process because H\&E is prone to produce different color intensities being conditioned by the temperature, storage conditions, and brand. Nevertheless, as stated by \citet{HI_Review}, HIs continue to be the gold standard for evaluating several types of tumors for cancer diagnosis. 

Computer-aided detection (CAD) and computer-assisted diagnostic (CADx) systems are continuously being developed to assist medical image analysis. Clinicians are heavily reliant upon CAD for cancer detection and monitoring. However, given the reliance on CAD and CADx for cancer detection, there is always an extra focus and need for developing systems that improve pathologists' productivity and ameliorate the reliance on outcomes by adding consistency to the diagnosis process and reducing observer subjectivity. Machine learning (ML) approaches are increasingly being used in CAD and HI analysis to diagnose cancer in various tissues or organs, such as the breast, prostate, skin, brain, bones, liver, and so on. Furthermore, when used in HI analysis, ML approaches reveal potential benefits. As a result, they have seen a lot of use in tasks like feature extraction, classification, and segmentation. The visual properties of macro vision images used in other machine learning applications, such as scene reconstruction, object and face recognition, differ from those of HIs, which contain complex textures and rich geometric structures. 

This paper proposes a novel method for texture characterization of HIs with a considerable success rate. We state that it is possible to quantify the intrinsic properties of HIs to the maximum extent by combining biodiversity and taxonomic measures and discrete wavelet transform. Thus, the main contributions of our research are: (i) an information-theoretical measure of ecological diversity indices and measures of biodiversity for texture characterization; (ii) the exploitation of independent wavelet subband coefficients' non-linear interactions across time; (iii) the mixture of wavelet features and statistical properties of taxonomic indexes representing an unexplored method based on a non-deterministic system analysis; (iv) such a mixture characterizes HIs so that intrinsic properties have provided promising performance for real-world HI datasets such as the CRC and the BreakHis datasets.  

This paper is organized as follows: Section \ref{sec:works} presents works that put forth descriptors more popular in medical image analysis. Section \ref{sec:method} puts forth the proposed approach as well as the concepts from which it originated. Section \ref{sec:results} presents the datasets, experiments results, and discussion. Finally, the last section presents the conclusion and perspectives of future work. 

\section{Related Works} \label{sec:works}
Several feature descriptors have been used to extract a relevant and discriminant information from images. Some are based on shape, texture, fractal, or combination of those mentioned above. Besides natural images, texture descriptors are becoming increasingly popular in medical image analysis, particularly in HIs due to the variability of texture that such images exhibit. Therefore, researchers have been studying a broad range of textural descriptors for the classification of HIs, which are expected to be invariant to translation, scale, rotation, and intensity changes. 

Characterizing morphological features from structures observed in HIs, as well as exploring higher-level representations capable of capturing relevant information for medical diagnosis purposes, is one of the most preeminent challenges in extracting features from such images. The above characteristics are related to recognizing tissue alterations (such as cell density or aberrant cell quantity) or cellular changes (e.g. malformed nuclei) caused by mitotic phases. Furthermore, morphological traits are related to how pathologists investigate HIs, looking for specific reasons to categorize them. High-level features, on the other hand, are generalizations of all structures in HIs, not solely cell structures. As a result, most researchers exploit representations or texture descriptors in the frequency domain \citep{HI_Review}. This section presents some state-of-the-art works for feature extraction from HIs. 
 
Several authors have utilized the descriptors based on the grey-level co-occurrence matrix (GLCM) to represent texture in HIs. \cite{Kuse2010235} employed GLCM for feature extraction with a segmentation process through unsupervised mean-shift clustering. The latter minimizes color variety to facilitate the segmentation of the image using thresholds. Following that, nuclei are identified, and overlapping is reduced employing contour and area constraints. \cite{Caicedo2011519} combined seven feature extraction methods, including GLCM, and constructed a kernel-based representation of the data for each feature type. The kernels are then used within an SVM to detect similarities between data for the implementation of a content retrieval mechanism. \cite{FernandezCarrobles201525}  proposed a feature extraction approach that uses frequency and spatial textons, implying that images are represented by a limited vocabulary of textures. Texton histograms and GLCM extracted from texton maps are used as classification features. Similarly, the effect of various colormaps on these procedures was assessed. Despite the fact that GLCM requires a gray-level image, the transformation of the H\&E color image to gray-level is influenced by the staining color variability, which affects GLCM accordingly. \cite{ISI:000391124500024} utilized a random forest to determine whether or not GLCM features are susceptible to image variations. The work also emphasized the significance of color normalization. 

\cite{Reis20172344} focused on stroma maturity to assess breast cancer. Image convolution with a bank of derivatives-of-Gaussian filters, basic image features, and LBP with multiple scales for the neighborhood produce the feature vector. \cite{ISI:000403573100015}  proposed geometric- and texture-aware features based on Hu moments and fractal dimensions, respectively. The latter was used to distinguish between mitotic and non-mitotic cells by detecting changes in geometrical and textural nuclei. \cite{CruzRoa201191} proposed a patching method on HI slides, intending to create small regions and extract scale-invariant feature transform (SIFT), luminance level, and discrete cosine transform features to generate a bag-of-words.

The local binary pattern (LBP) is one of the most commonly used texture descriptors. LBP was used in the context of multispectral HIs by \citet{Peyret201883}. An SVM was used for the evaluation, which aligns all spectra and uses pixels from all other bands. Similarly, it employs a multi-scale kernel size. When compared to the standard LBP and the concatenated spectra LBP, this feature extractor performed better. To deal with multi-scale HIs, \citet{TambascoBruno2016329}  used a curvelet transform. The LBP algorithm was used to extract features from curvelets coefficients, which were then reduced using ANOVA. \cite{7532841} proposed an algorithm that finds nuclei areas using adaptive and iterative thresholding and extracts texture information using histograms of oriented gradients and LBP. 

\cite{Huang2011579} proposed a two-step feature extraction method comprised of a receptive field for detecting regions of interest and sparse coding in their study. Sparse coding assembles features by combining patches from the same region. Furthermore, the mean and covariance matrix of receptive fields, as well as sparse coding, are utilized as final filters. \cite{Noroozi2016128}  proposed an approach for distinguishing basal cell carcinoma tumors from squamous cell carcinoma tumors in skin HIs by using Z-transform features derived from a mixture of Fourier transform features. \citet{Wan2017291} employed a dual-tree complex wavelet transform to represent the HIs in breast cancer detection for mitosis detection. The feature vector is made up of parameters from the symmetric alpha-stable distribution and the generalized Gaussian distribution. \citet{ISI:000391731800013}  used fractal dimension features to detect breast cancer. These features distinguish between malignant and benign tumors satisfactorily on HIs at 40$times$ magnification. Finally, \citet{ataky2021novel} presented a bio-inspired texture (BiT) descriptor based on biodiversity and taxonomic indices. The authors mapped breast cancer HI images to an abstraction model of an ecosystem, from which measures of species diversity, richness, evenness, and taxonomic distinctiveness were extracted. The resulting texture descriptor has been shown to be rotation, translation, and scale invariant. Experiments on HI datasets revealed that the BiT achieved competitive results when compared to deep methods, and outperformed traditional texture descriptors.

In the last years, deep features have become very popular in several image classification tasks, including HIs. \citet{KhalidKhanNiazi2016} put forward a CAD system for bladder cancer with a focus on extracting epithelium features with segmentation through an automatic color deconvolution matrix construction. \citet{spanhol2017} employed deep features extracted with a pre-trained AlexNet for the classification of breast tumors into benign and malignant. The method proposed by \citet{vo2019classification} put forth a feature extraction based on the combination of CNNs and boosting tree classifiers. The latter employs an ensemble of inception CNNs for visual features extraction from multi-scale images. Firstly, data augmentation methods were employed. Subsequently, ensembles of CNNs were trained to extract multi-context information from multi-scale images. The last-mentioned stage extracted both global and local features of breast cancer tumors. \citet{george2019deep} presented an approach for breast cancer diagnosis, which extracts features from nuclei based on CNNs. The methodology comprises different approaches for extracting nucleus features from HIs and selecting the most discriminative spatially sparse nucleus patches. A pre-trained set of CNNs was used to extract features from such patches. Afterward, features belonging to individual images are fused using 3-norm pooling to obtain image-level features. 

Various works employed and combined different categories of features for capturing information from geometrical structures and textures from HIs.
\citet{ISI:000391124500024} introduced a method aiming to quantify features' instability across a few prostate cancer datasets with known variations caused by staining, preparation, and scanning platforms. The author evaluated five groups of features, such as graph-based features, gland shape features, co-occurring gland tensor features, subgraph features, and Haralick texture features. \citet{ISI:000381691000001} investigated features that characterize lung cancer the best. They extracted the objective quantitative image features such as Haralick texture features of the nuclei, nuclei edge intensity, texture features of the cytoplasm, etc. 

The work of \citet{ISI:000256869500006} put forth a low-level to high-level mapping to facilitate image retrieval. Such mapping stage consists of color and gray-level histograms, Tamura texture histogram, LBP, Sobel histogram, and invariant feature histograms. \citet{Pang2017} raised a CAD system for lung cancer detection. Such a CAD system utilizes textural features, such as GLCM, Tamura, and LBP, and shape features, namely global features, SIFT, and morphological features. The work of \citet{Kruk2017357} employed textural, morphometric, and statistical (histogram) features to describe nuclei for clear-cell renal carcinoma grading. To this end, the Genetic algorithm and the Fisher discriminant were utilized to select essential features. \citet{6450064} introduced a multi-field-of-view classification method to determine low against high-grade ductal carcinoma from breast HIs. This method utilizes a multiple patch size procedure for WSI to analyze which textural, morphological, and graph-based features are the most relevant to each patch size. \citet{Tashk20156165} put forth a comprehensive framework for breast HI classification that evaluates mitotic pixels in L*a*b color space. A mixture of LBP, statistical features, and morphometrics is extracted from mitotic candidates. The work of \citet{CruzRoa201191} introduced a patching approach on HI slides to form small regions and extract SIFT, discrete cosine transform, and luminance level features to produce a bag-of-words. Besides, semantic features represent high-level information that can be associated with HIs to aid their classification.

\citet{5505922} compared four color spaces (RGB, L*a*b, gray-scale and RGB) with H\&E representation and eleven features such as GLCM, Zernike, Tamura, Chebychev, color histograms, Gabor, edge statistics, Chebyshev-Fourier, and others to represent lymph node HIs. \citet{De2013475} propose a fusion of several feature types for uterine cervical cancer HI classification. The authors employed a feature vector based on Delaunay triangulation, GLCM, and weighted density distribution. The work of \citet{Vanderbeck2014785} utilized textural, pixel neighboring statistics, and morphological features to represent seven categories of white regions of liver HIs. \citet{6868127} put forth a MIL method to detect Barrett's cancer from HIs. They utilized cell-level morphometric features, to wit, radius, perimeter, central power sums, area, the roundness of segments, and so forth, within regions and patch-level features such as color histograms, SIFT, LBP, and from segmented images employing the watershed algorithm. 

\citet{6943991,7367154} proposed a feature selection method of liver HI classification based on morphometric features and graph-based features. Pair of greedy algorithms such as fselector and in-house recursive was employed to select features in a collection of 200 features where an SVM classifier implemented the fitness function. The work of \citet{Michail20143374} highlighted nuclei using connected-component labeling to classify non-centroblast and centroblast cells. \citet{ISI:000403573100015} presented the so-called geometric- and texture-aware features based on Hu moments and fractal dimensional, respectively. The features set thereof was applied to detect geometrical and textural changes in nuclei to discriminate between non-mitotic and mitotic cells. The method introduced by \citet{Kong20091080} classifies neuroblastomas utilizing textural and morphological features. The authors consider that pathologists employ morphological features for their analysis, and textural features can be easily extracted. GLCM features and sequential floating forward selection were then utilized.

These are some examples of works developed to detect breast cancer through HIs. However, in spite of the substantial efforts and countless methods that have been proposed in recent years, accurate classification of HIs remains a challenge. Existing approaches developed to cope with such a purpose still arise performance issues in light of noise, different image resolutions, and a lack of a considerable amount of data. Furthermore, methods based on CNNs present a lack of explainability and interpretability, which are both necessary for understanding the behavior of such models, and trusting their decisions. Likewise, pre-trained CNN architectures designed for object classification still require fine-tuning some of their layers on a large amount of data to achieve good performance, including tiny architectures such as T-CNN Inception and T-CNN~\citep{MatosBOK19,Ataky2020}. 

\section{Proposed Approach} \label{sec:method}
This work proposes an efficient method for texture characterization of HIs combining ecology diversity measures and multi-resolution analysis. Combining biodiversity and taxonomic measures with the discrete wavelet transform (DWT) makes it possible to quantify the intrinsic properties of HIs to the maximum extent. Thus, the fundamental research highlights of the proposed method are: (i) an information-theoretical measure of ecological diversity indices and measures of biodiversity; (ii) the exploitation of non-linear interactions of single and independent wavelet subband coefficients over time; (iii) the mixture of wavelet features and statistical properties of taxonomic indexes representing an unexplored method for non-deterministic pattern's systems analysis. 

This section presents the concepts of information-theoretical measures of ecological diversity, multi-resolution analysis, and DTW, which are the base concepts of our method, before presenting our method for texture characterization of HIs. 


\subsection{Information-Theoretical Measure of Ecological Diversity}

The information-theoretical measures of ecological diversity are used in biology to compare behavioral patterns between species in different areas and within-neighborhood. Similarly, diversity indices based on species richness are of an underlying use when describing an all-inclusive behavior of an ecosystem, constituting a non-deterministic system of patterns.
\citet{ataky2021novel} proposed the bio-inspired texture descriptor (BiT) based on an information-theoretical measure of taxonomic diversity that includes the species' richness, abundances, and taxonomic distinctness. They consider an image as an abstract model of an ecosystem where pixels correspond to individuals, gray levels correspond to species, the number of different gray levels corresponds to species richness, and the number of distinct gray levels in a specific region corresponds to species abundance. 

Species richness of an image represents the number of gray levels therein contained. The higher the richness index, the more diverse the system is. The richness indices regard but the abundance of each species (gray levels) and among those present in BiT are Margalef’s diversity index (${D}_{{Mg}}$) and Menhinick’s diversity index (${D}_{{Mn}}$). Both are the ratio between the total number of gray levels recorded ($S$) and the total number of pixels in the image ($N$):

        \begin{equation}
            {D}_{{Mg}}= \frac{S-1}{\ln N}
            \label{eq:mg}
        \end{equation}
        
                 \begin{equation}
            {D}_{{Mn}}= \frac{S}{N}
            \label{eq:mn}
        \end{equation}
BiT integrates Shannon-Wiener ($d_{SW}$) and McIntosh’s ($e_M$) indices to account for the spatial variance of diversity. They are defined as the proportion of pixels of gray level $i$ in terms of $S$, and the ratio between the number of pixels in the $i$-th gray level and the total number of pixels, and the number of gray level in an image, respectively. 

\begin{equation}
{d}_{{SW}} = - \sum_{i=1}^{S}\left ( p_{i}\: \ln\: p_{i}\right )
    \label{eq:shannon}
\end{equation}
\begin{equation}
            e_M = \sqrt{\frac{\displaystyle\sum_{i = 1}^{S}{n_i^2}}{(N-S+1)^2 + S -1}}
            \label{eq:mci}
        \end{equation}
\noindent where $p_i$ denotes the proportion of pixels with the $i$-th gray level, and $n_i$ denotes the number of pixels of the $i$-th gray level (the summation is over all gray levels). 

Other diversity measures integrated into BiT are Berger-Parker dominance (${d}_{{BP}}$), which represents the ratio between the number of pixels in the most abundant gray level ($N_{max}$) and $N$, Fisher’s alpha diversity metric (${d}_{{F}}$), which represents the number of groups of closely related pixels, and Kempton-Taylor index of alpha diversity ($d_{KT}$), which quantifies the interquartile slope of the cumulative abundance curve. 
 
         \begin{equation}
            {d}_{{BP}}= \frac{N_{max}}{N}
            \label{eq:bpd}
        \end{equation}

\begin{equation}
            {d}_{{F}} = \alpha \ln\left(1 + \frac{N}{\alpha}\right)
            \label{eq:fam}
        \end{equation}

\begin{equation}
            {d}_{{KT}} = \frac{\displaystyle\frac{1}{2}n_{R_1} + \displaystyle \sum_{R_1+ 1}^{R_2 -1} n_r + \frac{1}{2}n_{R_2}}{\log \displaystyle\frac{R_2}{R_1}}
            \label{eq:kt}
        \end{equation}
        
\noindent where $\alpha$ is nearly equal to the number of gray levels represented by a single pixel, $n_r$ denotes the number of gray levels with abundance $R$, $R_1$ and $R_2$ are the 25\% and 75\% quartiles of the cumulative gray level curve, $n_{R_1}$ is the number of pixels in the class where $R_1$ falls, and $n_{R_2}$ is the number of pixels in the class where $R_2$ falls. 




Because such indices may be insensitive to taxonomic differences or similarities, \citet{ataky2021novel} have also integrated taxonomic indices, such as diversity and distinctness, which consider the taxonomic relationship between different pixels within an image. Taxonomic diversity ($\Delta$) includes aspects of taxonomic relatedness considering the abundance of different gray levels and the taxonomic relationship between them. Its value represents the average taxonomic distance between any two pixels, chosen randomly from an image. Taxonomic distinctiveness ($\Delta^*$), in turn, denotes a measure of pure taxonomic relatedness. It is the average taxonomic distance between two pixels of different gray levels. 

    \begin{equation}
        \Delta = \frac{\displaystyle \sum_{i=0}^{S} \sum_{i < j}^{S}w_{ij}x_{i}x_{j}}{\displaystyle\frac{N\left ( N-1 \right )}{2}}
        \label{eq:tdv}
    \end{equation}

    \begin{equation}
        \Delta^* = \frac{\displaystyle \sum_{i=0}^{S} \sum_{i < j}^{S}w_{ij}x_{i}x_{j}}{\displaystyle \sum \sum_{i< j}^{}x_{i}x_{j}}
        \label{eq:tdt}
    \end{equation}
    
\noindent where $x_i$ and $x_j$ denote the number of pixels that have the $i$-th and $j$-th gray level in the image, respectively, and $w_{ij}$ denotes the 'distinctness weight' (distance) given to the path length linking pixels $i$ and $j$ in the hierarchical classification. 

Other taxonomic indices integrated into BiT are the sum of phylogenetic distances (${s}_{PD}$), which denotes the sum of phylogenetic distances between pairs of gray levels, the average distance from the nearest neighbor ($ {d}_{NN}$), which denotes the average distance to the nearest group of one or more gray level of the image to form a unit, the extensive quadratic entropy (${e}_{EQ}$), which denotes the sum of the differences between gray levels, the intensive quadratic entropy (${e}_{IQ}$), which denotes the number of gray levels and their taxonomic relationships. It establishes a possible link between the diversity indices and the biodiversity measurement indices by expressing the average taxonomic distance between two randomly chosen gray levels. The relationships between the latter influence the entropy, in contrast to other diversity indices. Finally, the last taxonomic index is the total taxonomic distinctness (${d}_{TT}$), which denotes the average phylogenetic distinctiveness added across all gray levels. 

\begin{equation}
        {s}_{PD}  = \left ( \frac{S(S-1)}{2} \right )\frac{\displaystyle \sum \sum_{i < j^2}^{}ij^ai^aj}{\displaystyle \sum \sum_{i<j^a}i^aj^{}}
        \label{eq:spd}
\end{equation}
\begin{equation}
        {d}_{NN} = \sum_{i}^{S}\min\left ( d_{ij},a_i \right )
\end{equation}

\begin{equation}
{e}_{EQ} = \sum_{i\neq j}^{S} d_{ij}
\end{equation}

\begin{equation}
{e}_{IQ} = \frac{\displaystyle\sum_{i\neq j}^{S} d_{ij}}{S^{2}}
\end{equation}

\begin{equation}
{d}_{TT} = \sum i \frac{\displaystyle\sum_{i\neq j}^{S}d_{ij}}{S-1}
\label{eq:ttd}
\end{equation}

\noindent where $i$ and $j$ denote two distinct gray levels, and $a$ is the number of pixels that have such gray levels, and $d_{ij}$ is the distance between the gray levels $i$ and $j$; $a$ is the abundance of the referred gray level, and $S$ the total number of gray levels. 

Such ecological diversity measures may characterize the texture of HIs through its second-order statistical properties, which implies comparing neighboring pixels and defining how a pixel at a specific location relates statistically to pixels at different locations. Furthermore, because they rely on group analysis, they enable a behavioral exploration of the neighborhood of regions displaced from a reference location. In addition, the BiT descriptor takes advantage of the invariance characteristics of ecological patterns such as rotation, scale, and reflection. 

The BiT descriptor was developed to estimate diversity and parenthood between gray levels from images. The fundamental idea of combining biodiversity measures and taxonomic indices is to obtain a quantitative estimate of textural variability in space or time that can compare textural entities composed of several components. Applying these measures to HIs mapped as an ecosystem, they explore different periodicities and attempt to characterize their texture. This analysis is constrained to the neighboring of individual pixels, and the within-neighborhood periodicity properties can be used to determine texture differences between different regions. 

\subsection{Multi-Resolution Analysis}
The multi-resolution analysis is a signal processing strategy that employs filter banks to extract relevant information from signals, such as the frequencies and their locations depending on the duration of the signal, and at different resolutions ~\citep{castleman1996digital}. 
The motivation behind the multi-resolution analysis of HIs is to define each subband of the wavelet decomposition as a system modeled by ecological ecosystem diversity.

Like natural images, texture across HIs presents non-deterministic patterns that resemble a complex system. The wavelet decomposition plays a fundamental role in defining the spatial dependence structure (correlations between elements) because there is a direct analogy between wavelet subbands and physical systems of particles \citep{levadaComplexe}.
Considering that an image is modeled as an ecosystem, we exploit several wavelet subbands to reflect seasonal variation in the context of fluctuation-mediated coexistence. Such a variation in species may relate to the phenology, particularly in the context of season changes (time/frequency). Integrating statistical properties of taxonomic indexes from different variations (subbands) may help broader the view of ecosystems (intrinsic properties of an HI). Thus, with the wavelet decomposition, we exploit non-linear interactions of individual subband coefficients over time for capturing the textural information of HIs based on the principle that most ecosystems work in a cause-effect relationship. 

The succinct description of the multi-resolution analysis allows presenting two functions responsible for generating an entire wavelet system: the primary wavelet
and the scale function. 
The scale functions $\Phi_{q,r}$ and wavelets $\Psi_{p,q}$, are
orthogonal because they respect the following condition:

\begin{equation}
    \int_{-\infty }^{+\infty }{\Phi_{q, r}(x)\Psi_{q,r}(x)dx = 0 } 
    \label{eq:wavelet}
\end{equation}

\noindent where $q\in Z$ represents the scale of the function, and $r\in Z$ corresponds to the translation of $r/2^q$ concerning the scale function and the primary wavelet, given by $q=0$ and $r=0$. The scale function and the wavelet are defined in $\mathbb{R}$.
The translation parameter corresponds to the time information in the transform domain, and the scaling parameter represents the signal compression and expansion process~\citep{mallat1999wavelet}. 
\begin{figure}[htpb!]
	\centering
    \includegraphics[width=3.5in]{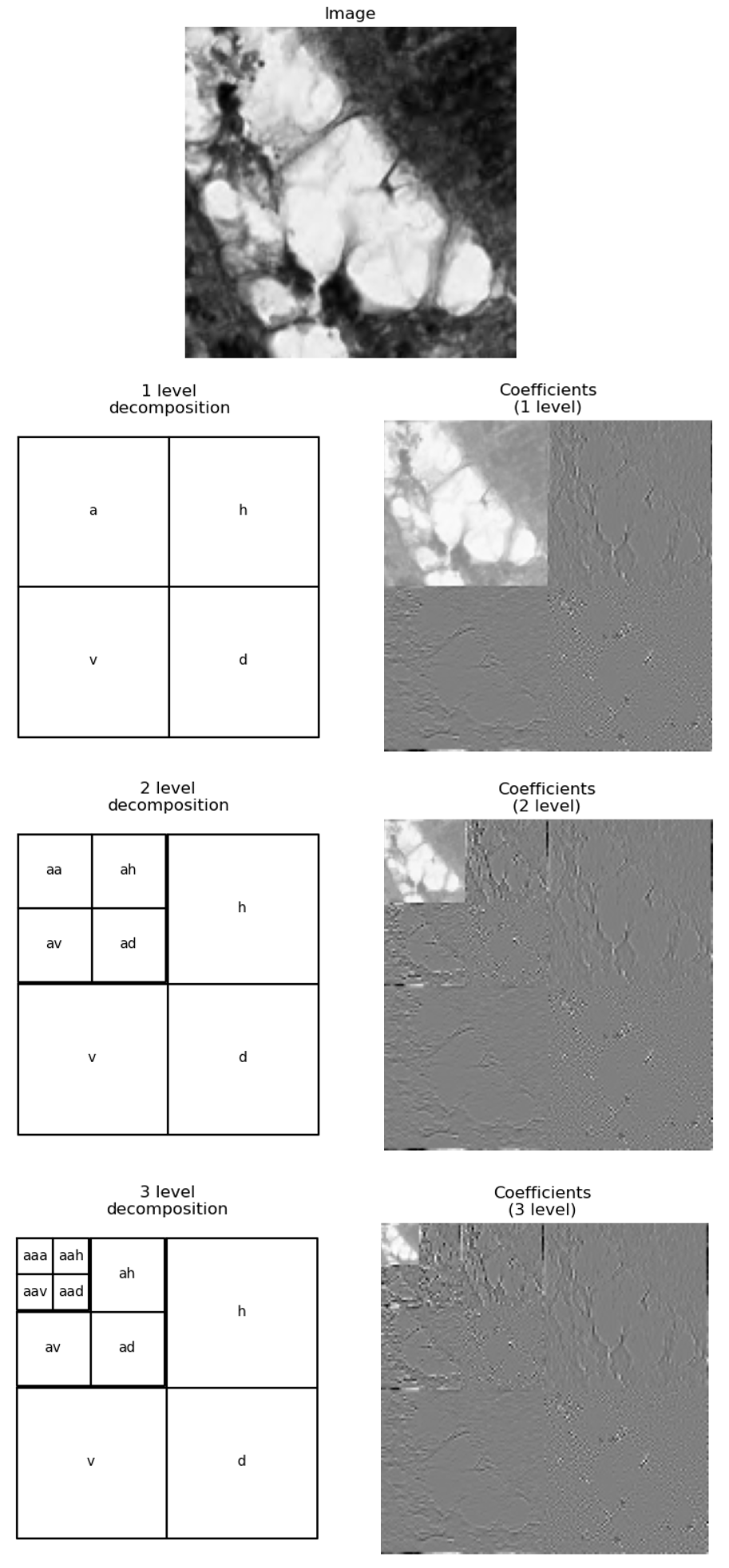}
	\caption{Multi-resolution representation of an image.} 
    \label{fig:wavelet1}
\end{figure}

Fig.~\ref{fig:wavelet1} illustrates the decomposition of an HI into three levels through the wavelet transform. The useful information obtained from the initial image of the vessel, on a smaller scale, is found in the six adjacent squares. The information contained in these squares is called detail or resolution, which is the information needed to move from one degree of refinement (or "sharpness") to another. By adding the information related to the squares, it is possible to recompose the image. This way of decomposing and recomposing images can be implemented quickly and effectively, employing wavelet transforms. 

The continuous wavelet transform can be expressed as:
\begin{equation}
    CWT(\tau,a)= \int_{-\infty }^{+\infty }f(t)\frac{1}{\sqrt{a}}\Psi^{*} \left(\frac{t-1}{s} \right)dt
    \label{eq:cwt}
\end{equation}

\noindent where $\tau$ and $a$ represent the translation and scale parameters, respectively.
However, this transformation requires infinite translations and scaling. 


In the discrete wavelet transform (DWT),
wavelets are not scaled or translated continuously but at discrete intervals, which is achieved by modifying the continuous wavelet as: 

\begin{equation}
    \Psi_{s,\tau}(t) = \frac{1}{\sqrt{\left | s \right |}}\Psi \left ( \frac{t-\tau}{a} \right )
    \label{eq:dwt1}
\end{equation}
\begin{equation}
    \Psi_{q,r}(t) = \frac{1}{\sqrt{\left | s_{0}^{q} \right |}}\Psi \left ( \frac{t-r\tau_{0}s_{0}^{q}}{s_{0}^{q}} \right )
    \label{eq:dwt2}
\end{equation}

\noindent where $q$ and $r$ are integers, $s_0>$1 is a fixed expansion parameter, and $\tau_0$ is the translation factor that depends on the expansion factor. Generally, $s_0=$2 is chosen for a frequency sampling called dyadic sampling, and $\tau_0=$1 is chosen for temporal sampling, also dyadic~\citep{oliveira2007analise}. This results in: 

\begin{equation}
    \Psi_{q, r}(t) = \sqrt{2^{q}}\Psi(2^{q} - r)
    \label{eq:dwt3}
\end{equation}



A wavelet can be seen as a low-pass filter, and a series of scaled wavelets can be seen as a bank of band-pass filters with a $Q$ factor (filter bank fidelity factor).
DWT can be applied to decompose an HI into two other HIs through low-pass filters $l$ (scaling signals) and high-pass filters $h$ (wavelet signals). Therefore, DWT can be constructed as a perfect reconstruction filter bank with pairwise quadrature mirror filters $l$ and $h$~\citep{Nsimba2019}. The synthesis filters $l$ and $h$, used to reconstruct the original signal from wavelet coefficients, are required for a complete DWT specification. The inverse discrete wavelet transform is the term for this technique. 

\begin{figure}[htpb!]
	\centering
    \includegraphics[width=2.8in]{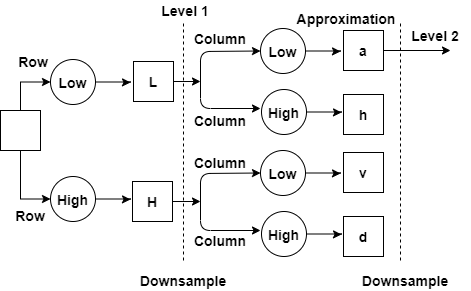}
	\caption{Wavelet decomposition of an HI.} 
    \label{fig:wavelet2}
\end{figure}

Fig.~\ref{fig:wavelet2} shows the decomposition of an HI, and it consists of four subbands: approximation (a), horizontal detail (h), vertical detail (v), and diagonal detail (d). In the next scale, the subband image (a) is used for DWT computation.
\subsection{Ecological Modeling of Wavelet Subbands}
\begin{figure}[htpb!]
    \includegraphics[width=0.45\textwidth]{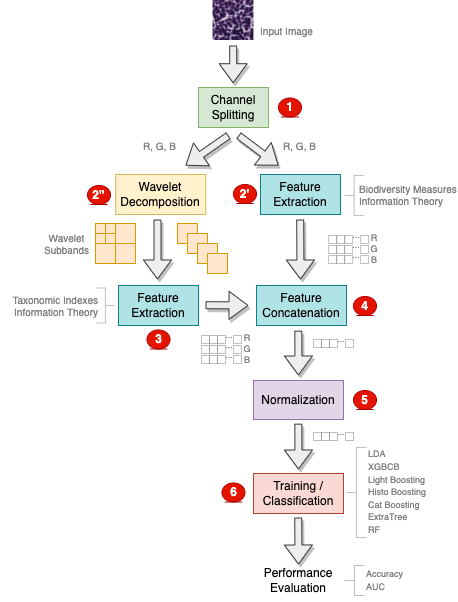}
\caption[Methodology]{General overview of the proposed scheme. }
	\label{Fig:model}
\end{figure}

The proposed method is employed and integrated for texture classification as follows: (1) image channel splitting; (2$^\prime$) wavelet subband decomposition; (2$^{\prime\prime}$) computation of biodiversity measures and information theory from each channel (R, G, and B) to form a features vector; (3) computation of taxonomic indexes and information theory from each wavelet subband to form a features vector; (4) feature vector concatenation; and (5) classification and performance evaluation. Fig.~\ref{Fig:model} shows the general overview of the proposed scheme.

 \paragraph{Channel Splitting:} we applied the integrative method to make each image channel (R, G, B) a separate input. The motivation is to exploit color information. Therefore, we represent and characterize an input image by a set of local descriptors generated from the interaction of a pixel with its neighborhood from a given channel (R, G, or B) as well as wavelet subbands thereof. 

 \paragraph{Feature Extraction:} after the channel splitting stage, each image channel undergoes feature extraction. Intrinsic properties and discriminant characteristics within each input channel are extracted as follows. First, for each channel (R, B, and G), we perform a multi-resolution analysis of a texture employing a wavelet transform, generating four subbands: \textit{a, h, v}, and \textit{d}. Since we used three levels, the subband \textit{a} is used for DWT computation in the next scale. After, we compute taxonomic measures, Shannon entropy, and total information, resulting in a 9-dimensional vector for each subband. This sequence can be seen in Fig.~\ref{Fig:model} (steps 2$^\prime$ and 3). Taxonomic measures comprise taxonomic diversity, taxonomic distinctness, the sum of phylogenetic distances, average distance from the nearest neighbor, extensive quadratic entropy, intensive quadratic entropy, and total taxonomic distinctness \citep{ataky2021novel}. Because we have split an input image into three channels and chosen a 3-level wavelet decomposition (leading to 10 subbands), step 3 will produce a 270-dimensional feature vector (9$\times$3$\times$10). In parallel, we compute biodiversity measures and again Shannon entropy and total information directly from each original image channel resulting in a 9-dimensional vector for each channel. Biodiversity measures from step 2$^{ \prime\prime}$ comprise Margalef’s and Menhinick’s diversity indexes, Berger-Parker dominance, Fisher’s alpha diversity metric, Kempton-Taylor index of alpha diversity, McIntosh’s evenness measure, and Shannon-Wiener diversity index \citep{ataky2021novel}. The reason for not employing DWT in this stage is that low-pass decomposition filters may present negative coefficients in the subband \textit{a} depending on filter coefficients used for decomposition. Nevertheless, biodiversity measures such as abundance and richness are non-negative. Because we have split an input image into three channels, step 2$^{\prime\prime}$ will produce a 27-dimensional feature vector (9$\times$3). Finally, feature vectors resulting from steps 2$^{ \prime\prime}$ and 3 are concatenated to form the final feature vector (step 4 in Fig.~\ref{Fig:model}). We named it BiTW descriptor because it results from the concatenation of biodiversity measures information theory, and taxonomic indexes extracted from wavelet subbands. The BiTW is a 297-dimensional feature vector, which may leave a whole path to a possible feature selection step. However, this is out of the scope of this paper. 

\paragraph{Normalization:} the feature vectors are split into training and test sets before the training step. Then, normalization and scaling occur independently on each feature in the training set, where values are normalized to the range $[0,1]$ using the min-max normalization. Minimum and maximum are then stored to be used on feature normalization over the testing data. The same procedure is used for the $k$-fold cross-validation (CV), where feature vectors are split into $k$ folds and computing the min-max pairs in the merged training folds. The min-max pairs obtained on the trained data are employed to normalize the training and the test folds This procedure is repeated for each new training/test fold during the CV procedure. 
 
 \paragraph{Training and Classification:} after feature extraction, the resulting feature vectors are taken through the classification process utilizing seven classifiers: histogram-based algorithm for building gradient boosting ensembles of decision trees (HistoB), light gradient boosting decision trees (LightB), fast, scalable, high-performance gradient boosting decision trees (CatBoost), extra trees (ExtraT), random forest (RF), gradient boosting decision trees (GB), and linear discriminant analysis (LDA). The performance analysis is conducted afterward using accuracy and area under the receiver operating characteristic curve (AUC). 

\section{Experiments and Results} \label{sec:results}
This section presents the two datasets used to assess the performance of the BiTW descriptor and the experimental protocol to evaluate the properties of the BiTW descriptor and its performance on classification tasks considering different classification and ensemble methods. We compare the BiTW descriptor's performance with shallow and deep state-of-the-art approaches.

\subsection{Image Datasets}
Two medical datasets were used in the experiments:

\begin{itemize}
    \item \textbf{CRC} \citep{CRC_1}(Fig.~\ref{Fig:crc}): is a dataset of colorectal cancer histopathology images of 5,000$\times$5,000 pixels that were patched into 150$\times$150 images and labeled according to the structure they contain. Eight types of structures are labeled: stroma (ST), tumor (T), complex stroma (C), immune or lymphoid cell (L), mucosa (M), debris (D), adipose (AD), and empty or Background (E). There is a total of 625 images per structure type, resulting in 5,000 images.
    
    \begin{figure}[htpb!]
    	\centering
    	\includegraphics[width=0.35\textwidth]{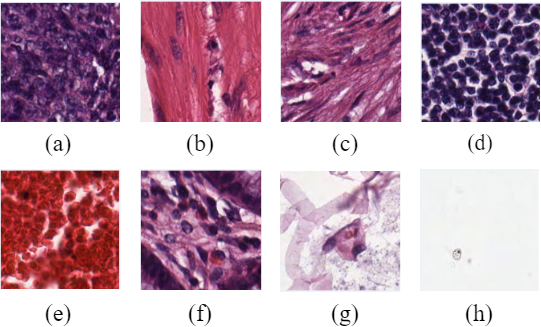}
    	\caption[CRC Dataset]{Example of HIs from the CRC dataset: (a) Tumor, (b) Stroma, (c) Complex, (d) Lympho, (e) Debris, (f) Mucosa, (g) Adipose, (h) Empty.}
    	\label{Fig:crc}
    \end{figure}
    
    \item \textbf{BreakHis} \citep{Spanhol2016} (Fig.~\ref{Fig:breakhis}): comprises 9,109 microscopic images of breast tumor tissue collected from 82 patients using different magnification factors (40$\times$, 100$\times$, 200$\times$, and 400$\times$). To date, it contains 2,480  benign and 5,429 malignant samples (700$\times$ 460 pixels, 3-channel RGB, 8-bit depth in each channel, PNG format). The dataset is imbalanced.
    
    \begin{figure}[htpb!]
    	\centering
    	\includegraphics[width=0.38\textwidth]{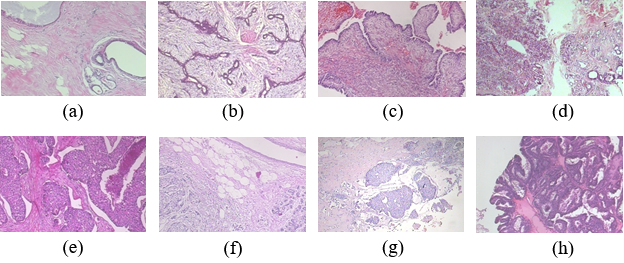}
    	\caption[BreakHis dataset]{Example of HIs from the BreakHis dataset: (a) Adenosis, (b) Fibroadenoma, (c) Phyllodes, (d) Tabular adenomaa, (e) Ductal carcinoma, (f) Lobular carcinoma, (g) Mucinous carcinoma, (h) Papillary carcinoma, where (a-d) are benign tumors and (e-f) malignant tumors.}
    	\label{Fig:breakhis}
    \end{figure}
\end{itemize}

\subsection{Experimental Results}
Table~\ref{tab:acc_crc} presents the accuracy of both the monolithic classifiers and ensemble methods trained with feature vectors from BiTW on the CRC dataset. GB yielded the best results for both train-test split and $k$-fold CV in the group of all classification algorithms. Likewise, we have computed the area under the ROC curve (AUC), another important metric commonly used in medical images, which accounts for the true-positive rate against the false-positive rate at different threshold levels. The AUC achieved by the BiTW description with GB was 0.99. It is worthy of note that AUCs of 0.7 to 0.8 are considered acceptable, 0.8 to 0.9 are considered excellent, and greater than 0.9 is deemed to be exceptional. 

\begin{table}[!htpb]
\setlength{\tabcolsep}{4pt}
\caption{Accuracy (\%) and AUC of monolithic classifiers and ensemble methods with the BiTW descriptor on the CRC dataset for train-test split and 10-fold CV.}
\label{tab:acc_crc}
\footnotesize
\centering
 \begin{tabular}{l c c c c c c c} 
 \toprule
\multirow{2}{*}{\parbox{1.5cm}{\centering Experimental Protocol (Metric)}}         & \multicolumn{7}{c}{Classification Algorithms}\\ 
\cmidrule{2-8}     
& HistoB & LightB & LDA & CatB & ExtraT & RF & GB\\  
  \midrule
70/30 (Acc) & 92.42 & 92.20 & 89.27 & 92.20 & 92.42 & 92.42 & \textbf{93.28} \\
70/30 (AUC) & 0.991 & 0.992 & 0.988 & 0.993 & 0.993  & 0.991 & 0.991 \\
\midrule
10-fold (Acc) & 91.12 & 91.81 & 89.45 & 91.33 &{91.53}&90.59 & \textbf{93.73}\\
10-fold (SD)  & $\pm$0.05 & $\pm$0.01 & $\pm$0.08 &$\pm$0.02 & $\pm$0.03 & $\pm$0.02& $\pm$0.02 \\
10-fold (AUC) & 0.990 & 0.992 & 0.989 & 0.993 & 0.993 & 0.991 & 0.994 \\
 \bottomrule
  \multicolumn{8}{l}{Acc: Accuracy; SD: Standard Deviation.}
\end{tabular}
\end{table}

Table~\ref{tab:acc_crc_relatedWorks} compares the results achieved by BiTW + GB with the state-of-the-art for the CRC dataset. The BiTW slightly outperforms the accuracy achieved by almost all other methods. For instance, the difference in accuracy to the second-best method (shallow) is 0.77\%, and with the third-best method (CNN) is 1.33\%, considering an 8-class classification task and 10-fold CV. For 5-fold CV, BiTW slightly outperformed the second-best method (CNN) with the difference of 1.0\%. It is noteworthy to highlight that CNNs generally require massive labeled datasets and, when not possible, may need pre-trained models and/or data augmentation to learn high-quality representations. In the medical field, however, this is not always possible. The shallow approaches using the BiTW descriptor, unlike the latter, did not require any data augmentation on the CRC dataset and has proven to be promising relative to CNNs as well as other shallow methods, despite HIs having other structures than textures.

\begin{table}[!htpb]
\setlength{\tabcolsep}{2pt}
\caption{Average accuracy (\%) of shallow and deep approaches on the CRC dataset for 5-fold CV, 10-fold CV, and AUC.}
\label{tab:acc_crc_relatedWorks}
\footnotesize
\centering
 \begin{tabular}{l c l c c c} 
  \toprule
      \multirow{2}{*}{Reference} & \multirow{2}{*}{Approach} & \multicolumn{2}{c}{Accuracy (\%)} &  & \multirow{2}{*}{AUC}\\
  \cmidrule{3-4}
            &          & 10-fold & 5-fold &  & \\ 
  \midrule
 \citet{CRC_ribeiro2019classification} & Shallow & 97.60$^*$ & -- &  & \bf{0.994} \\
 \citet{CRC_sarkar2017sdl}& Shallow & 73.60 &  -- &  & --\\
  \citet{crc_jorgensen2017}& Shallow & -- &  -- & & 0.960\\
   \citet{CRC_kather2016multi} & Shallow & 96.90$^*$ & -- &  & --\\
 \citet{CRC_kather2016multi} & Shallow & 87.40 & -- &  & --\\
 \citet{crc_naiyar}& Shallow & -- &  -- &  & 0.960\\
  \citet{crc_rathore}& Shallow & -- &  -- &  & 0.970\\
  \citet{crc_kalkan}& Shallow & -- &  -- &  & 0.950\\
 \citet{crc_Masood}& Shallow & -- &  -- & & 0.900\\
  \citet{ataky2021novel} & Shallow &  92.96 & -- &  & -- \\
 {\bf BiTW+GB} & Shallow & \bf 93.73 & \textbf{93.60 $\pm$0.2} &  & \bf{0.994} \\
 \citet{CRC_wang2017histopathological} & CNN & -- & {92.60} &  & --\\
 \citet{CRC_pham2017scaling} & CNN & -- & 84.00 & & --\\
 \citet{CRC_rkaczkowski2019ara} & CNN & 92.40 & 92.20 &  & --\\
 \bottomrule
 \multicolumn{4}{l}{$^*$Used 2-classes classification instead (malignant and benign).}
\end{tabular}
\end{table}

Table~\ref{tab:acc_breakHis} presents the accuracy of both the monolithic classifiers and ensemble methods trained with BiTW feature vectors on the BreakHis dataset. By employing the train-test split, the LightB achieved the best accuracy of 99.26\% and 98.62\% for 40$\times$ and 200$\times$ magnifications, respectively. The ExtraT achieved 98.50\% and HistoB 98.38\% accuracy for 100$\times$ and 400$\times$ magnifications, respectively. Furthermore, the AUC is nearly 0.98 or above regardless of the classifier or ensemble method. Considering the $k$-fold CV, ExtraT yielded the best accuracy of 98.75\% and 98.63\% for 40$\times$ and 100$\times$ magnifications, respectively. The LightB, in turn, achieved an accuracy of 98.72\% for 200$\times$ magnification. Finally, HistoB yielded an accuracy of 98.38\% for 400$\times$ magnification. 
\textcolor{black}{Furthermore, we carried out experiments with $k$-fold cross-validation to ensure that every sample from the original dataset has the chance of appearing in the training and test set, which is a best practice in the presence of limited data. Table \ref{tab:acc_breakHisFolds} presents the average accuracy with the BiTW descriptor on the BreakHis dataset at image level with a 10-fold CV. The results for both types of dataset splitting are very similar for nearly all the classifiers and ensemble methods.}

\begin{table}[!htpb]
\setlength{\tabcolsep}{2pt}
\caption{Accuracy (\%) and AUC of monolithic classifiers and ensemble methods with the BiTW descriptor on the BreakHis dataset at image level with train-test split.}
\label{tab:acc_breakHis}
\footnotesize
\centering
 \begin{tabular}{c c c c c c c c c} 
 \toprule
 \multirow{2}{*}{\parbox{1.7cm}{\centering Image Magnification}}        & \multicolumn{8}{c}{Classification Algorithm}\\ 
\cmidrule{2-9}
    & Metric & HistoB & LightB & LDA & CatB & ExtraT & RF & GB\\  
\midrule
\multirow{2}{*}{40$\times$}     &Acc & 98.97 & \textbf{99.26} & \textbf{99.26} & 98.25 & 98.97 & 98.25 & 98.18 \\
     &AUC& 0.992  & {0.995} & {0.993} & 0.981 & 0.985 & 0.988 & 0.987 \\
\multirow{2}{*}{100$\times$} & Acc    & 98.43 & 98.30 & 92.41 & 97.19 & \textbf{98.50} & 97.19 & 96.64 \\
      &AUC& 0.991  & {0.990} & {0.982} & 0.987 & 0.992 & 0.987 & 0.980 \\
\multirow{2}{*}{200$\times$}       & Acc& 98.55 & \textbf{98.62} & 90.88 & \textbf{98.62} & 98.55 & 98.00 & 97.58 \\
       &AUC& 0.989  & {0.993} & {0.963} & 0.994 & 0.992 & 0.989 & 0.988 \\
\multirow{2}{*}{400$\times$}      &  Acc  & \textbf{98.38} & 97.98 & 90.29 & 97.74 & 97.98 & 97.19 & 97.74 \\
      &AUC& 0.993  & {0.990} & {0.983} & 0.987 & 0.989 & 0.982 & 0.986 \\
 \bottomrule
\end{tabular}
\end{table}

\begin{table}[!htpb]
\setlength{\tabcolsep}{2pt}
\caption{Average accuracy (\%) of monolithic classifiers and ensemble methods with the BiTW descriptor on the BreakHis dataset at image level with 10-fold CV.}
\label{tab:acc_breakHisFolds}
\footnotesize
\centering
 \begin{tabular}{c c c c c c c c} 
 \toprule
    \multirow{2}{*}{\parbox{1.7cm}{\centering Image Magnification}}     & \multicolumn{7}{c}{Classification Algorithm}\\ 
\cmidrule{2-8}
      & HistoB & LightB & LDA & CatB & ExtraT & RF & GB\\  
  \midrule
40$\times$&98.62 & 98.61 & 98.61 & 98.06 & \textbf{98.75} & 98.04 & 98.12  \\
100$\times$ & 98.48 & 98.45 & 91.60 & 97.74 & \textbf{98.63} & 97.87 & 97.76 \\
200$\times$  & 98.54 & \textbf{98.72} & 91.55 & 97.98 & 98.65 & 97.98 & 98.53 \\
400$\times$   & \textbf{98.76} & 98.74 & 91.00 & 98.08 & 98.33 & 97.58 & 97.77 \\
\bottomrule
\end{tabular}
\end{table}

Table~\ref{tab:acc_breakhis_relatedWorks} compares the results achieved by the proposed approach with the state-of-the-art for the BreakHis dataset. 
\textcolor{black}{Though different classifiers outperformed others for different magnifications, for the sake of fairness, we choose a classifier to present for comparison with related work based on the average performance of the four magnifications. Computing the average we obtain 98.58\%, 98.54\%, 93.21\%, 97.95\%, 98.50\%, 97.65\% and 97.53\%, for HistoB, LightB, LDA, CatB, ExtraT, RF and GB. Thus, HistoB is chosen for comparison purposes because it presented the highest average accuracy.
The BiTW descriptor with HistoB achieved a substantial accuracy of 98.97\%, 98.43\%, 98.55\% and 98.38\% for 40$\times$, 100$\times$, 200$\times$, and 400$\times$ magnifications, respectively. It is important to notice that the proposed approach outperforms the accuracy of shallow and deep methods, regardless of the magnification. The differences in accuracy between the proposed method and the second- and third-best methods are 1.47\% (Shallow) and 1.97\% (CNN), 0.93\% (CNN) and 1.63\% (Shallow), 1.35\% (CNN) and 2.75\% (Shallow), 1.18\% and 3.18\% (Shallow) for 40$\times$, 100$\times$, 200$\times$, and 400$\times$ magnifications, respectively.} 

\begin{table}[!htpb]
\setlength{\tabcolsep}{2pt}
\caption{Average accuracy (\%) of shallow and deep approaches on the BreakHis dataset. All these works used the same data partitions for training and test.}
\label{tab:acc_breakhis_relatedWorks}
\footnotesize
\centering
 \begin{tabular}{l c c c c c c} 
 \toprule
          &  & \multicolumn{4}{c}{Image Magnification}\\ 
\cmidrule{3-6}
  Reference & Method & 40$\times$ & 100$\times$ & 200$\times$ & 400$\times$ & \\
 \midrule
 \citet{BrealHis_2}& CNN &  97.00 & { 97.50} & {97.20} & { 97.20} &  \\
 \citet{BreakHis_3}& CNN &  92.80 & 93.90  & 93.70 & 92.90 & \\
 \citet{BreakHis_4}& CNN &  83.00 & 83.10  & 84.60 & 82.10 & \\
 \citet{BreakHis_5}& CNN &  90.00 & 88.40  & 84.60 & 86.10 & \\
 \citet{BreakHis_7}& CNN &  94.10 & 93.20  & 94.70 & 93.50 & \\
 \citet{BreakHis_8}& CNN & 88.23  & 84.64 & 83.31 & 8.98 &\\
 \citet{BreakHis_9}& CNN & -- & 95.00 & -- & -- &  \\
 \citet{Spanhol2016}$^{*}$ & Shallow &  75.60 & 73.00  & 72.90 & 71.20 &\\
 \citet{Spanhol2016}$^{+}$ & Shallow &74.70 & 76.80 & 83.40 & 81.70 & \\
\citet{BreakHis_1}$^{*}$& Shallow &  88.30 & 88.30  & 87.10 & 83.40 & \\ 
\citet{ataky2021novel}$^{\dagger}$& Shallow   &  { 97.50} & 96.80  & 95.80 & 95.20\\
{\bf BiTW + HistoB}& Shallow   &  {\bf 98.97} & {\bf 98.43}  & {\bf98.55 }& {\bf 98.38} & \\
\bottomrule
\multicolumn{6}{l}{ $^{*}$LBP descriptor; $^{+}$GLCM descriptor;$^{\dagger}$BiT descriptor.}
\end{tabular}
\end{table}

\subsection{Discussion}
The proposed approach was assessed with two HI datasets, both with eight classes. The experiment protocol employed a train-test split (70/30) and a $k$-fold CV. For either experimental protocol, the results led to the following findings:

\begin{itemize}
    \item[(1)] Exploiting information-theoretical measures of ecological diversity indices in conjunction with non-linear interactions of single and independent wavelet subband coefficients throughout time yielded promising results.
    \item[(2)] Although HIs contain other structures than texture, it was possible to characterize texture and achieve a good discriminating capability by employing biodiversity measures and taxonomic indexes together with multi-resolution analysis through DWT.
    \item[(3)] Such a mixture allowed the characterization of HIs to such an extent that intrinsic properties have provided a promising performance for the real-world datasets classification, reaching 93.73\% accuracy and 0.994 AUC for the CRC dataset. Regarding the results on the BreakHis dataset, the accuracy were 99.26\%, 98.50\%, 98.62\%, and 98.76\% for 40$\times$, 100$\times$, 200$\times$ and 400$\times$ magnification, respectively. The AUC of all the magnifications was above 0.98.
\end{itemize}
Overall, the proposed approach outperformed state-of-the-art shallow and deep approaches on CRC and BreakHis datasets, regardless of the non-textural information that HIs may contain.

\section{Conclusion}
\label{sec:conclusion}
The current research leveraged the information-theoretical measure of ecological diversity indices together with a discrete wavelet transform to characterize texture across HIs. We explored the interactions of individual wavelet subband coefficients over time and modeled each as an ecosystem from which measures of biodiversity and statistical properties of taxonomic indexes are extracted to represent HI texture effectively. We stated that by combining measurements of biodiversity from each HI channel and taxonomic indexes extracted from different wavelet subbands, it should be possible to quantify the intrinsic properties of such images to the maximum extent. The mixture of wavelet features and statistical properties of ecology diversity indexes represent a novel method and a promising tool for quantifying intrinsic properties of texture across HIs. Wherefore, the experimental results have shown an increase in terms of texture discrimination over both HI datasets. Moreover, the proposed method outperformed several shallow and deep state-of-the-art methods.

In future work, we intend to further improve the classification accuracy by exploring various color spaces and different families of wavelets to find an optimized wavelet that will bring on discriminative information for characterizing HI categories. Furthermore, because an increasing number of DWT-level decompositions can result in high-dimensional feature vectors, we want to look into appropriate feature selection approaches or dimensionality reduction techniques to find feature subspaces that not only improve texture features but also improve the proposed method's class discrimination capability.

\section*{Acknowledgment}
This work was funded by the Regroupement Strategique REPARTI - Fonds de Recherche du Québec - Nature et Technologie (FRQNT) and by the Natural Sciences and Engineering Research Council of Canada (NSERC) under Grant RGPIN 2016-04855. 

\bibliographystyle{model2-names.bst}\biboptions{authoryear}
\bibliography{refs}

\section*{Supplementary Material}
Supplementary material that may be helpful in the review process should be prepared and provided as a separate electronic file. That file can then be transformed into PDF format and submitted along with the manuscript and graphic files to the appropriate editorial office.

\end{document}